\documentclass[sigconf]{acmart}

\usepackage{booktabs} 
\usepackage{amsmath,bm}
\usepackage{enumitem}
\usepackage{footnote}
\setcopyright{rightsretained}
\usepackage{threeparttable}
\usepackage{amsmath, amssymb, amsfonts, amsthm}

\acmDOI{10.475/123_4}

\acmISBN{123-4567-24-567/08/06}

\acmConference[LAK'19]{9th International Conference on Learning Analytics and Knowledge}{March 4-8, 2019}{Tempe, Arizona USA}
\acmYear{2019}
\copyrightyear{2019}

\acmArticle{4}
\acmPrice{15.00}

\begin{document}
\title{Goal-based Course Recommendation}

\begin{abstract}
 With cross-disciplinary academic interests increasing and academic advising resources over capacity, the importance of exploring data-assisted methods to support student decision making has never been higher. We build on the findings and methodologies of a quickly developing literature around prediction and recommendation in higher education and develop a novel recurrent neural network-based recommendation system for suggesting courses to help students prepare for target courses of interest, personalized to their estimated prior knowledge background and zone of proximal development. We validate the model using tests of grade prediction and the ability to recover prerequisite relationships articulated by the university. In the third validation, we run the fully personalized recommendation for students the semester before taking a historically difficult course and observe differential overlap with our would-be suggestions. While not proof of causal effectiveness, these three evaluation perspectives on the performance of the goal-based model build confidence and bring us one step closer to deployment of this personalized course preparation affordance in the wild.
\end{abstract}

\author{Weijie Jiang}
\affiliation{
  \institution{University of California, Berkeley \& Tsinghua University}
}
\email{jiangwj@berkeley.edu}

\author{Zachary A. Pardos}
\affiliation{
  \institution{University of California, Berkeley}
}
\email{pardos@berkeley.edu}

\author{Qiang Wei}
\affiliation{
  \institution{Tsinghua University}
}
\email{weiq@sem.tsinghua.edu.cn}


%
%



\maketitle

\section{Introduction}
The terrain of a university degree program can be difficult to successfully traverse. Challenging decisions abound, such as which major to declare, subject(s) to explore, and level of difficulty of course load to take on. These decisions involve hard to balance risk vs. reward trade-offs, made more difficult by the multiple objectives students want to maximize and risks they want to hedge against (e.g., choosing challenging courses of value to employers while maintaining high GPA). Given the abundance of historic data on student enrollments, grades, and majors, a question naturally arises if learning analytics approaches can extract any wisdom from these records that may aid students in achieving their goals. In this paper, we build on the findings and methodologies of a quickly developing literature around prediction and recommendation in higher education to introduce an approach to goal-based course recommendation. 

As multidisciplinary educational interests increase, such as with data science, so too does the importance of providing appropriate intellectual on-ramps to subject matter that promotes equity and inclusion in these pursuits. This means providing pathways to success for students from various disciplinary backgrounds. We focus on this particular goal of finding appropriate preparation course(s), given one's existing curricular exposure, for a target course of interest. There are a variety of reasons why existing prerequisite course information provided by the university may not be satisfactory: (1) the prerequisites may not be up to date (2) they may not be comprehensive, neglecting to include combinations of courses from different departments that together would cover the requirement material (3) they do not take into account what an individual student already knows, and are thus often bypassed by students if not enforced, and (4) they may consist of often oversubscribed courses for which students may have no choice but to seek alternatives for. Our proposed approach addresses these four potential shortcomings, especially by tailoring suggestions of the preparatory class based on a model of the knowledge a student has already acquired. 

The task of recommending a set of appropriate courses personalized to any student's course history and any arbitrary target course is arguably of intractable difficulty for one person. Faculty tend to be local experts with deep knowledge within their subject area. Non-faculty academic advisers have broader course familiarity, but at the expense of depth, and both resources are scarce in higher education compared to the number of students enrolled. Machine learning models can scale and benefit from the breadth and depth of representations learned from big data but lack the ability to easily tease apart the difference between correlation and causation based on observations. We explore if, given enough constraints, reasonable suggestions can be reliably extracted from such a model. We choose three prediction validations (grade prediction, prerequisite prediction, and course selection prediction) intended to, collectively, suggest if the approach may warrant testing in the wild. We choose Recurrent Neural Networks (RNN) as the framework to extend to this goal-based recommendation task due to their robust representational and temporal capabilities. While RNNs have been previously applied to make recommendations based on collaborative filtering principles \cite{hidasi2015session, hidasi2016parallel, pardos2018connectionist}, they have not been re-purposed to make more targeted personalized goal-based recommendations in any domain. Our validation and application of RNN typologies to a goal-based task is thus a novel contribution of the work.

\section{Related Work}
Recent findings suggest that models capturing co-enrollment information surpass those only using features of students or of courses in capturing variation in student performance \cite{Gardner:2018:CNR:3170358.3170373}. Part of the benefit of co-enrollment information in predicting performance stems from capturing the interaction effect that can occur when a student takes two difficult courses at once \cite{Brown:2018:CCA:3170358.3170366}. Other feature engineering approaches to course grade prediction found that better discrimination between grade labels could be achieved by binarizing the ordinal grade label \cite{polyzoufeature}. This same work found that student features better predicted high grades compared to course features, which predicted the low grade class, though low grades remained difficult to predict in part due to being the minority class. Features of student and course proved useful to the task of elective course selection prediction but were substantially outperformed when feature sets of student-course ratings could be taken into account \cite{estebanhybrid}. Classical collaborative filtering was used to predict course grades, which served as weights on an inferred pre-requisite graph meant to potentially help students towards higher rates of on-time graduation \cite{backenkohlerdata}. Though high accuracy was achieved, their context was limited, using only 72 courses from a single department. Work focusing on prediction of on-time graduation has found neural embeddings of students from enrollment sequences to be accurate, particularly after a student's second year of study \cite{luo2018diagnosing}.

The use of an RNN to predict student course grades can be seen as a type of assessment model. RNNs have been used for assessment in the educational contexts of games, to predict outcomes based on game activity \cite{akramimproving}, and to predict responses to questions of various skills given response histories in math tutoring systems \cite{piech2015deep}. Their inferences have been used to produce pre-requisite graphs in the same math tutoring contexts as well as in MOOCs \cite{chenbehavioral}. They have also been used as models of behavior prediction, to suggest the next resource a learner is likely to spend time on next \cite{pardos2017enabling}. In the context of higher education, they have been used in a deployed campus system to predict the next courses a student is likely to enroll in, given their course taking history \cite{pardos2018connectionist}. 

The intended or realized operationalization of grade prediction models have primarily been applied to early-warning type systems meant to signal to students or advisers when struggle is occurring or is imminent \cite{Hlosta:2017:OEI:3027385.3027449,zhang2018early}. Early-warning implementations like this can experience unintended consequences, however, such as leading to greater course drop-out \cite{jayaprakash2014early}. One system attempted a more pre-emptive approach, showing grade distributions of courses to students and common course sequences for each course before enrollment; however, an evaluation of the system found that course selection behaviors were not affected by the system but GPA was, leading to an unexpected quarter of a grade point \textit{decrease} in GPA \cite{chaturapruek2018data}. These findings underscore the daunting task of achieving meaningful academic improvement as a result of analytic intervention; however, a more specific lesson can be learned. Both instances of intervention underachievement have in common grade related data being shown directly to students. This may have the effect of signaling that pathways are closed off or that a particular achievement can be expected regardless of effort. We believe it is therefore important for analytic-based interventions to encourage learners to set goals and for these interventions to strive to be more prescriptive in scaffolding avenues to achieving them.

While on-campus course analytics have focused on between-course enrollment data, tangential research in MOOCs has focused on models of predicting course outcomes based on within-course activities \cite{Andres:2018:SMC:3170358.3170369, gardner2018student, gardner2018evaluating, le2018communication}. It should be noted that even given a positive validation and eventual real-world beneficial effect of our approach, it would not be a panacea for student success. There is no shortage of dimensions to the story of student achievement in higher education. Work showing the correlation between course outcomes and timely access to course materials among late enrolling students \cite{Agnihotri:2017:ISC:3027385.3027437} serves as a reminder of this.

\section{Goal-based recommendation approach}
We build our approach around several assumptions. The first is that students have a zone of proximal development \cite{chaiklin2003zone} with respect to course material and that course recommendations should be limited to courses they are expected to be able to succeed in. This necessitates a predictive model of course grades to be trained, akin to the Deep Knowledge Tracing neural framework \cite{piech2015deep} applied to tutoring system. The second assumption is that such a model of course performance is capable of inferring prerequisite information that can subsequently be used to recommend courses anticipated to be appropriate preparation for a target course. To validate this assumption, we use the university's existing prerequisite courses list and test the grade prediction model's ability to infer these existing dependencies. Lastly, we assume that the recommendations generated by our model ought to be followed more frequently by students who succeed in a target course than students who underachieve. This assumption gives way to our third validation, predicting the previous semester course enrollments before a historically difficult course in the next semester. A relevant example of correlation not equalling causation is the case of students who take an honors course being likely to do well in courses in the subsequent semester, not because of the intrinsic preparatory value of the honors course but because the self-selected students who take them are generally high achieving. We acknowledge this confound but believe that this validation, paired with the first assumption, of not recommending courses to students they are unlikely to pass, mitigates the concern. Furthermore, we limit recommendation to courses that are not of a higher level than the target course, according to the three division levels indicated by the course number (i.e., lower division, upper division, and graduate). We also constrain the recommendations to departments that contain prerequisite courses to other courses found in the target course's department. We hypothesize that these constraints mitigate the chances of an egregiously poor recommendation being made due to confounds in the data. 

Traditional Recurrent Neural Networks (RNNs) have been used to predict the next action in a sequence. This amounts to a collaborative recommendation of the nature of "most people like you did X next." When it comes to students' diverse intentions in selecting courses, a student's goal may not align with what most people have done. A simple approach could be to only train on students who achieved the intended goal; however, this approach is unsatisfying as it would eliminate data points that could be used to learn more robust representations of the domain. It is also undesirable as it would require that thousands of independent models be trained to correspond to our task of arbitrary target course preparation.

\subsection{Model Definition}
A canonical RNN takes as input a sequence of vectors $x_1,..., x_T$, mapped to a predicted output sequence of vectors $y_1,...,y_T$. This is achieved by computing a sequence of `hidden` states $h_1,...,h_T$, which can be viewed as successive encodings of relevant information from past observations that will be useful for future predictions. 

We use a popular variant of RNNs called Long Short-Term Memory (LSTM) \cite{hochreiter1997long}, which helps RNNs learn temporal dependencies with the addition of several gates which can retain and forget select information \footnote{RNNs were evaluated; however, consistently underperformed LSTMs}. They have been shown to generalize using long sequences more effectively than simple RNNs  \cite{bengio2003neural, gers1999learning} by using a gated structure to mitigate the vanishing gradient phenomenon. While our sequences are not long, our prediction task can benefit from the ability to selectively treat certain course grades as irrelevant (i.e. forgettable) to the prediction of future course grades. The decision of what information to forget at time step t is made by the forget gate $f_t$, where $\boldsymbol{g}_t$ is the course grades representation at time step t and $\boldsymbol{h}_{t-1}$ is the hidden state of the network at time step t-1.
\begin{equation*}
f_t = \sigma (\boldsymbol{W}_{fg}\boldsymbol{g}_t + \boldsymbol{W}_{fh}\boldsymbol{h}_{t-1} + \boldsymbol{b}_f)
\end{equation*}
The decision of which information will be stored into the cell is determined by calculating the input gate $i_t$ and a candidate value $\widetilde{\boldsymbol{C}_t}$ to be added to the state.
\begin{equation*}
i_t = \sigma(\boldsymbol{W}_{ig}\boldsymbol{g}_t + \boldsymbol{W}_{ih}\boldsymbol{h}_{t-1} + \boldsymbol{b}_i)
\end{equation*}
\begin{equation*}
\widetilde{\boldsymbol{C}_t} = {\rm tanh}(\boldsymbol{W}_{Cg}\boldsymbol{g}_t + \boldsymbol{W}_{Ch}\boldsymbol{h}_{t-1} + \boldsymbol{b}_C)
\end{equation*}
We denote $\boldsymbol{C}_t$ as the internal cell state and update it by the intermediate results calculated below. 
\begin{equation*}
\boldsymbol{C}_t = f_t \times \boldsymbol{C}_{t-1} + i_t \times \widetilde{\boldsymbol{C}_t}
\end{equation*}
Finally, we calculate LSTM's hidden state of time step t, denoted as $\boldsymbol{h}_t$.
\begin{equation*}
o_t = \sigma(\boldsymbol{W}_{og}\boldsymbol{g}_t + \boldsymbol{W}_{oh}\boldsymbol{h}_{t-1} + \boldsymbol{b}_o)
\end{equation*}
\begin{equation*}
\boldsymbol{h}_t = o_t \times {\rm tanh}(\boldsymbol{C}_t)
\end{equation*}

\begin{figure}
\includegraphics[height=1.2in, width=3.5in]{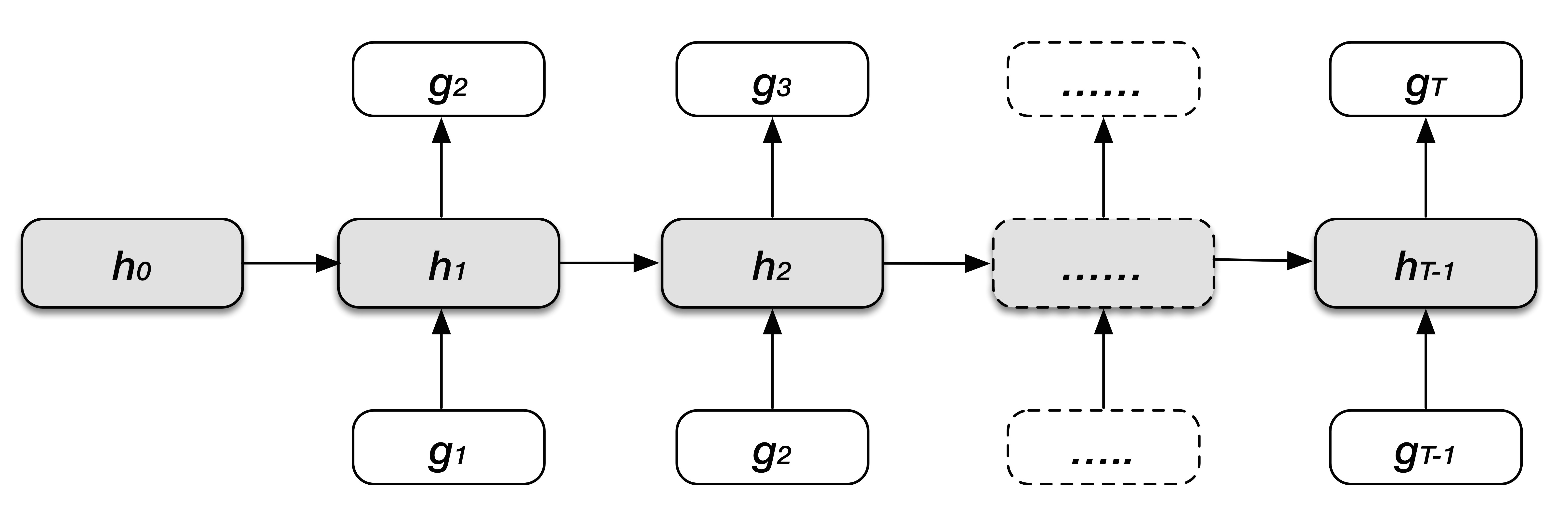}
\caption{Model 1 - Simple course grade prediction model}
\label{rnn1}
\end{figure}

Intuitively, a simple LSTM can be applied to the grade prediction task, where the output in each time slice is a vector representing the probability of receiving a certain grade for each of courses in the next time slice (semester), as is shown Figure \ref{rnn1}, referred to as \textit{Model 1} in the rest of the paper. 

\begin{figure}
\includegraphics[height=1.3in, width=3.5in]{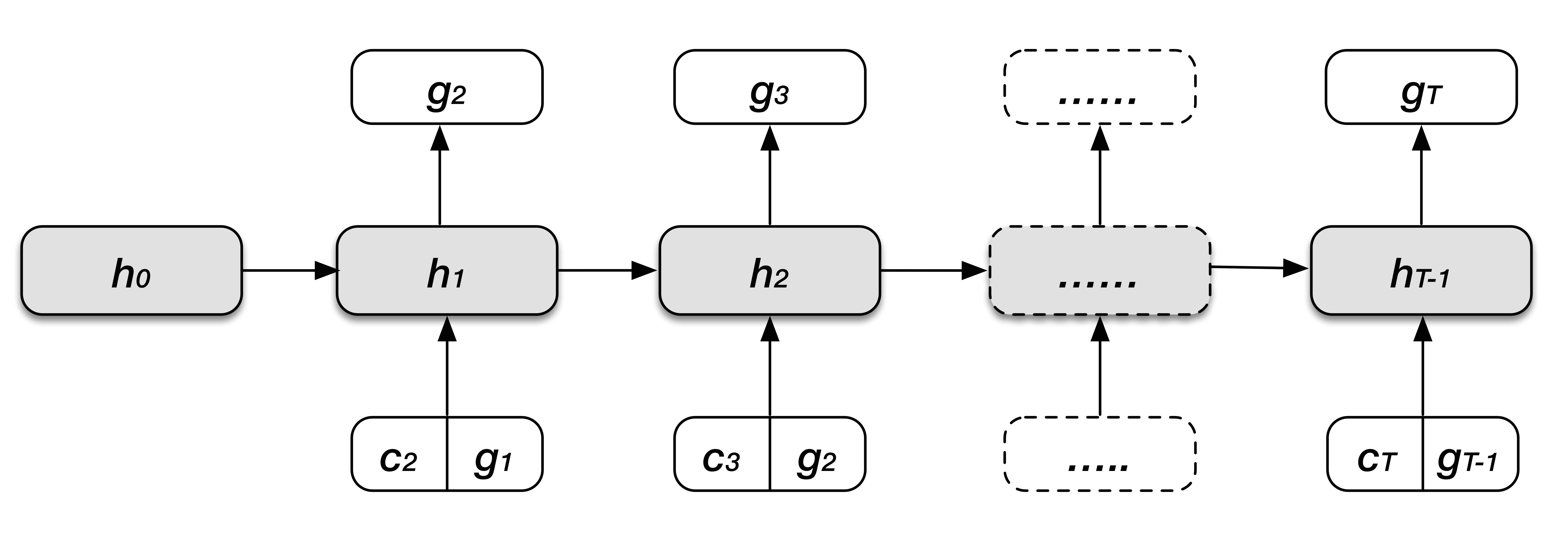}
\caption{Model 2 - Course grade prediction model with previous semester course grades and current semester co-enrollment as input to the hidden layer}
\label{rnn2}
\end{figure}

However, recent findings suggest that not only students' grades in previous semesters will influence their grades in the current semester, but also the course co-enrollment composition in the current semester will impact their performance \cite{Brown:2018:CCA:3170358.3170366}. The reasons lie in interaction effects among courses enrolled in together, such as (1) the zero sum of student available time and the time demands of each enrolled course and (2) positive synergistic effective among courses, for example, learning \textit{Data Structures} and \textit{Discrete Math} together may reinforce learning between courses because they share similar content. Hence, we present a variation on the simple LSTM which concatenates a multi-hot of courses co-enrolled in for the current semester $t+1$ (without grades) to the input of course grades from the previous semester $t$ as the input, aiming at predicting grades for semester $t+1$. See Figure \ref{rnn2} for illustration of the model, referred to as \textit{Model 2} in the rest of the paper. 

\begin{figure}
\includegraphics[height=2.2in, width=3.5in]{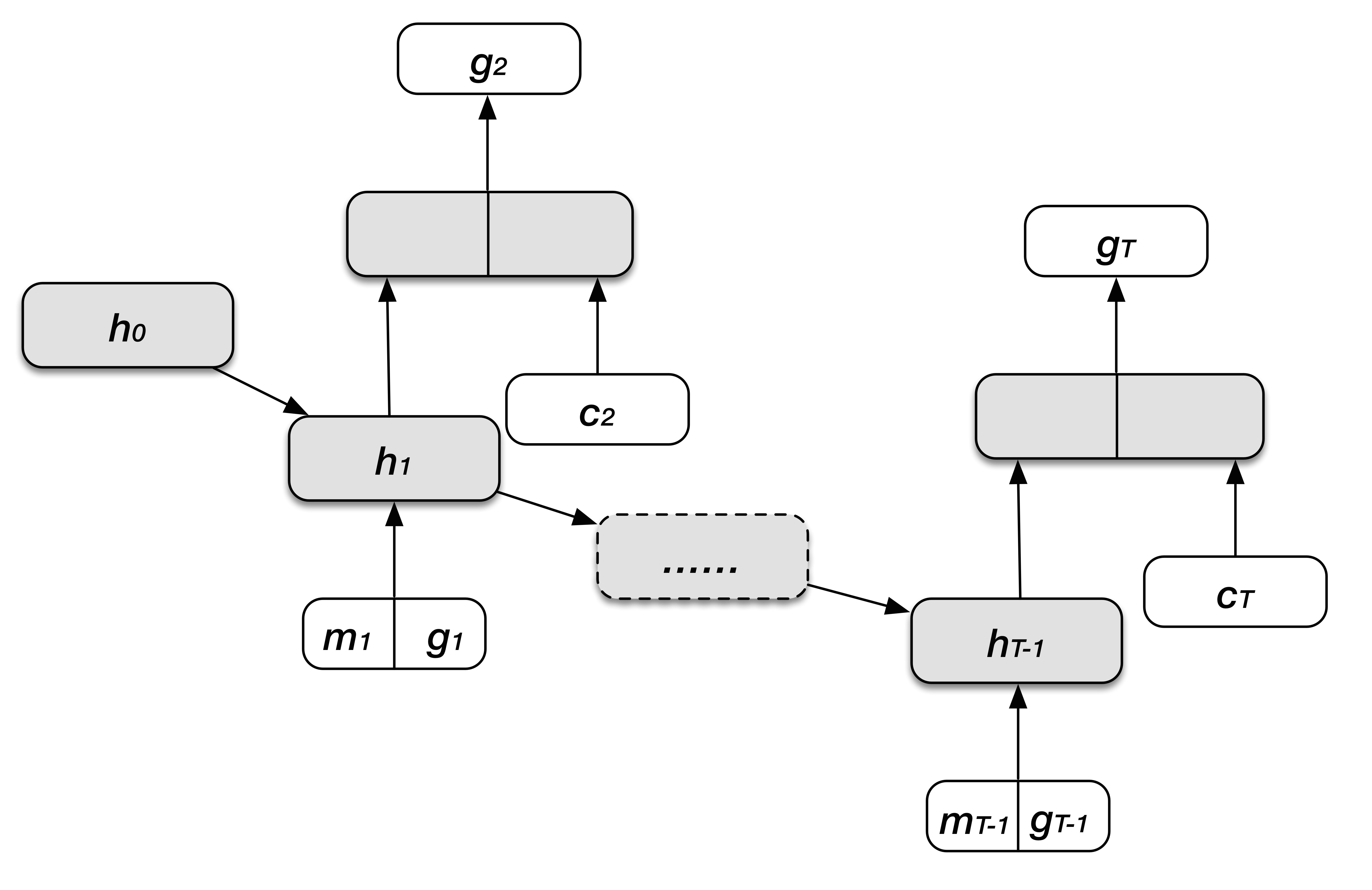}
\caption{Model 3 - Course grade prediction model with previous semester course grades, previous semester declared major(s), and current semester co-enrollment as input directly to the output layer}
\label{rnn3}
\end{figure}

Moreover, we assume that student major may affect the expected grade distributions of courses. For example, it may be expected that students achieve higher or lower grades in courses outside their major. Therefore, we present another variant of LSTM which concatenates major to grades in semester $t$. We feed the multi-hot of courses co-enrolled to a linear layer concatenated with the hidden layer. This means courses co-enrolled information in semester $t+1$ will only influence that semester without influencing all the hidden states and outputs in the following time slices\footnote{We made this choice after observing worse performance on a development set of a model which fed the courses co-enrolled information to the hidden layer}. See Figure \ref{rnn3} for illustration of the model, referred to as \textit{Model 3} in the rest of the paper.

\subsection{Input and Output Time Series}
In order to train an LSTM on student enrollment grades sequences, it is necessary to convert an enrollment grades sequence into a sequence of fixed length input vectors $g_t$. Note that a student can select several courses in a semester and get either a categorical grade, e.g., A, B, C, D, or a binary grade, i.e., Pass and No-pass for each enrolled course. Hence, the input should be  designed to consider all the information in the enrollment sequences. Assume that there are $n$ courses and $m$ categorical grades for each course in total. Let $g_t$ represent the grades of a student for all the courses enrolled in semester $t$, and specifically, $g_t^i$ to be a student's grade in semester $t$ for course $i$. Therefore, $g_t$ is set to be a multi-hot encoding to represent the combination of which courses were enrolled and which grades were received for those courses, and $g_t^i$ is set to be a one-hot encoding to represent the grade for course $i$ in semester $t$.

\begin{equation}
    \bm{g}_t^i = (s_i^1, s_i^2, ..., s_i^m, s_i^{Pass}, s_i^{No-Pass})
\end{equation}
and
\begin{equation}
    \bm{g}_t = (\bm{g}_t^1, \bm{g}_t^2, ..., \bm{g}_t^n) 
\end{equation}
So $\bm{g}_t^i \in \{0, 1\}^{m+2}$ and $\bm{g}_t \in \{0, 1\}^{(m+2)*n}$. 

We set $c_t$ to be a multi-hot encoding of multiple courses enrolled in semester $t$, where

\begin{equation}
    \bm{c}_t = (c_t^1, c_t^2, ..., c_i^n)
\end{equation}

Additionally, a student can have multiple majors in a semester. Therefore, we set $\bm{m}_t$ to be another multi-hot encoding of a student's majors in semester $t$, 

\begin{equation}
    \bm{m}_t = (m_t^1, m_t^2, ..., m_i^k)
\end{equation}
where $k$ is the number of all possible majors.

For the models proposed in Section 3.1, inputs are different concatenations of $\bm{g}_t$, $\bm{c}_t$ and $\bm{m}_t$, and outputs are always $\bm{g}_t$, meaning that we only predict student grades in next semester with variations of input feeding to LSTM, as are shown in Figures \ref{rnn1}, \ref{rnn2} and \ref{rnn3}. It is worth mentioning that in Figure \ref{rnn1}, the output in a certain time slice $t+1$ is predicting the conditional probability of course grades in semester $t+1$ given all the historical course grades of a student, i.e., $P(\bm{g}_{t+1}|\bm{g}_{1}, ..., \bm{g}_{t})$. In Figure \ref{rnn2}, the output in a certain time slice $t+1$ is predicting the conditional probability of course grades in semester $t+1$ given all the historical enrolled courses and their grades of a student , i.e., $P(\bm{g}_{t+1}|\bm{g}_{1}\bm{c}_{2}, ..., \bm{g}_{t}\bm{c}_{t+1})$. Lastly, in Figure \ref{rnn3}, the output in a certain time slice $t+1$ is predicting the conditional probability of course grades in semester $t+1$ given all the historical course grades of a student, the courses taken in semester $t+1$, and all the student's historical majors, i.e., $P(\bm{g}_{t+1}|\bm{g}_{1}\bm{c}_{2}\bm{m}_1, ..., \bm{g}_{t}\bm{c}_{t+1}\bm{m}_{t})$.

\subsection{Custom Masked Loss Function for Optimization}
Only grade labels for the courses a student enrolls in and completes can be used to calculate the loss. Therefore, the predictions of courses not taken in the next semester must be masked so as not to affect the loss calculation. The training objective is the negative log-likelihood of the observed sequence of student grades under the model. Let $\hat{\bm{g}}_{t+1}$ be the multi-hot encoding of which grade is received for which courses in semester $t+1$, which is the label for our prediction, normally, cross entropy loss will be applied to maximize the similarity between the distributions of the output of softmax layer and the label.

\begin{equation}
   loss = - \sum_t \hat{\bm{g}}_{t+1}^T \textrm{log} \bm{g}_{t+1}
   \label{loss}
\end{equation}

However, as mentioned, in the course enrollment grade prediction scenario, the course grades in a semester should not affect the grades of the courses not enrolled in for the next semester. For the grade of a course, $\bm{g}_t^i$, a student can receive only letter grade or Pass/No-Pass grade, i.e., one-hot in either $(s_i^1, s_i^2, ...,s_i^n)$ or $(s_i^{Pass}, s_i^{No-Pass})$. Therefore, two modifications to the loss function in Equation \ref{loss} are designed to deal with the two points above.

\begin{description}[font=$\bullet$~\normalfont\color{black}]
\item [\textbf{Not-enrolled in Courses Grade (first level) Mask}]
Grades for not-enrolled in courses in semester $t+1$ according to the labels, $\hat{\bm{g}}_{t+1}$, are masked in the loss function. 
\item [\textbf{Unrelated Grade Type (second level) Mask}] Within the grade encoding for each course, $\bm{g}_t^i$, we employ two separated cross-entropy loss functions for $\bm{g}_t^{i1} = (s_i^1, s_i^2, ...,s_i^n)$ and $\bm{g}_t^{i2} = (s_i^{Pass}, s_i^{No-Pass})$ and mask the unrelated one.
\end{description}

\begin{figure}
\includegraphics[height=1.0in, width=3.5in]{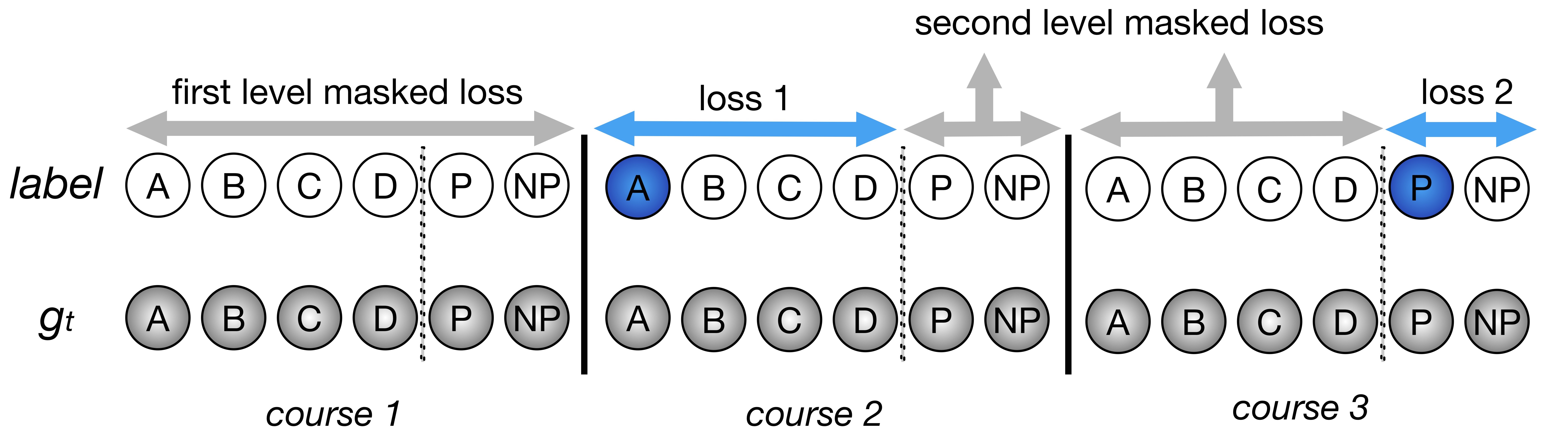}
\caption{Illustration for the two-level masked loss architecture}
\label{mask}
\end{figure}

The architecture of the two-level masked loss is illustrated in Figure \ref{mask}, which covers all the cases in the masked loss function. Specifically, Figure \ref{mask} assumes three courses in which \textit{course 2} and \textit{course 3} are enrolled in and with A and Pass received, respectively, according to the labels where a blue circle represents the real grade. Therefore, according to the not-enrolled in courses grade (first level) mask, all the outputs related to \textit{course 1} should be masked according to the unrelated grade type (second level) mask, Pass/No-Pass part and letter grade part of \textit{course 2} and \textit{course 3} in the outputs should be masked, respectively. The first level and second level masked losses are not calculated and back-propagated while only loss 1 and loss 2 are calculated and back-propagated through the network to update the weights of the model.

The overall modified loss function is naturally expressed as: 

\begin{equation}
   loss = - \sum_t \sum_{i, \hat{\bm{g}}_{t+1}^i\neq \bm{0}} (\hat{\bm{g}^{i1}}_{t+1}^T \textrm{log} \bm{g}_{t+1}^{i1}+\hat{\bm{g}^{i2}}_{t+1}^T \textrm{log} \bm{g}_{t+1}^{i2})
   \label{loss1}
\end{equation}

\section{Dataset}

We used a dataset from University of California, Berkeley, which contained anonymized student course enrollments from Fall 2008 through Spring 2017. The dataset consisted of per-semester course enrollment information for 164,196 students (both undergraduates and graduates) with a total of 4.8 million enrollments. A course enrollment meant that the student was still enrolled in the course at the conclusion of the semester. The median course load during students' active semesters was four. There were 10,430 unique courses, including 9,714 unique primary lecture courses from 197 subjects in 124 different departments hosted in 17 different divisions of 6 colleges. In all analyses in this paper, we only considered primary courses (lecture) and courses with at least 20 enrollments total over the 10 year period.
The raw data were provided in CSV format by the university's Enterprise Data Warehouse. Each row of the course enrollment data contained semester and grade information, an anonymous student ID, entry type (transfer student or new freshman), and declared major(s) at each semester. Course information included course name, course number, department, instructor, subject, enrollment count, and capacity. The basic structure of the enrollment data is shown in Table \ref{data}. 

\begin{table}[]
\centering
\caption{Example student enrollments from our dataset}
\label{data}
\begin{tabular}{p{1.8cm}p{1.2cm}p{0.8cm}p{0.8cm}p{1cm}p{0.8cm}l}
\toprule

 \textbf{Semester Year} & \textbf{STU ID (anon)} &
 \textbf{Major} &
  \textbf{Dept} &  \textbf{Course Num} & \textbf{Grade}  \\ \hline
Spring 2014 & x137905  & Law & Law & 178 & B  \\ 
Summer 2014 & x137905  & Law  & Law & 165 & C  \\ 
Fall 2014 & x282243 & Math & Math & 140 & D  \\ 
Fall 2014 & x282243 & Math  & Math & 121 & A  \\ 
\bottomrule
\end{tabular}
\end{table}

\section{Student Grade prediction}
In this section, we explain how we train the models proposed in Section 3.1 to predict course grades. Based on the sequential information in our data set, we separated the data set into three parts by time for evaluation, i.e., data from Fall 2008 to Fall 2015 as training set, data in Spring 2016 as validation set and data in Spring 2017 as test set. 

The loss function in Equation \ref{loss1} was minimized using stochastic gradient descent on minibatches. To prevent overfitting during training, dropout was applied to the last linear layer to compute the output. Weight decay and gradient clipping were also applied to prevent overfitting. We tuned the model on the dimensionality of the hidden layer and found 50 to be consistently the best for all the models. We fixed the mini-batch size to 32. Notably, in the goal-based recommendation scenario, we choose to allow for students to set the achievement level desired in the target course to either the A or B level. We refined the input and output of our model by setting this threshold (A or B), and then converted the categorical grades to binary classes, i.e., `above or equal to grade threshold' and `below threshold'. In all the following experiments, we tried both A and B as thresholds\footnote{We published our code for the paper regarding the model, and three validations, at https://github.com/CAHLR/goal-based-recommendation}. 

\subsection{Results}

\begin{table}
  \centering
  \caption{Evaluation of student course grade prediction}
  \label{grade}
    \begin{tabular}{c|c|c|c|c}
    \toprule
    \multicolumn{2}{c|}{\textbf{model settings}} & \multicolumn{2}{c|}{\textbf{letter grade}} &
    \multicolumn{1}{c}{\textbf{pass/no-pass}} 
    \\
    \midrule
    \textbf{threshold} & \textbf{model} & \textbf{accuracy} & \textbf{F-score} & \textbf{accuracy}  \\
    \midrule
    B     & Baseline-B & 85.46 & None & 90.97  \\
    \midrule
     B     & Model 1 & 87.75 & 39.05 & 90.71  \\
    \midrule
    B     & Model 2 & 88.05 &  42.01  & 91.78 \\
    \midrule
    B     & Model 3 & 87.78 & 40.21  & 91.76  \\
    \midrule
    A    & Baseline-A & 50.31 & None  & 83.69   \\
    \midrule
    A     & Model 1 & 74.61 & 55.05  & 85.42  \\
    \midrule
    A     & Model 2 & 75.23 & 60.24 &  85.81  \\
    \midrule
    A     & Model 3 & 75.19 & 58.86  &  86.05  \\
    \bottomrule
    \end{tabular}%
  \label{tab:grade}%
  \vspace{-4mm} 
\end{table}%

 We applied \textit{accuracy} and \textit{F-score} for the two binary classification tasks, i.e., letter grade classification and Pass/No-pass grade classification, which are shown in Table \ref{tab:grade}. The column `letter grade' shows the classification accuracy of `above or equal to grade threshold' for the letter grade type, while the column `pass' represents the classification accuracy of pass grade for the Pass/No-pass grade type. The column `F-score' is for the letter grade classification. The results show modest prediction performance. Current semester co-enrollment information was useful (Model 2 vs. Model 1) in the case of both threshold models. Major (Model 3) was not useful in either A or B threshold models compared to without (Model 2). The reason may lie in that the differences among students' majors may be already embedded in their various enrollment patterns, which means major information is not able to further boost the model's discriminative power in grade prediction. While the B threshold models performed better in terms of accuracy, they were also much closer to the performance of the majority class baseline (88.05 vs. 85.46). It will be tested in the next section if this close proximity to baseline performance prohibits the model from containing valid prerequisite relationships in its course embedding. The A model performed better than the B model in terms of F-score (60.24 vs. 42.01) and in terms of gain over baseline. In the case of the A model, major information was able to improve predictions in Pass grade, whereas major may have led to overfitting in the B model, given the strong majority class. In the following sections, we only evaluate these best models depicted in Figures \ref{rnn2} and \ref{rnn3}.

\section{Prerequisite Courses Prediction}
Instructor specification of prerequisites for his or her course is meant to ensure students have the necessary foundation and experience to be able to learn and succeed in the course. In order for our grade prediction based model to have utility in recommending preparation courses, it ought to encode the prerequisite relations between course pairs already specified by some courses. We designed a technique, inspired from \cite{piech2015deep}, to explore those pairs based on the learned model, which does not need any other model training. 

\begin{figure}
\includegraphics[height=2in, width=3in]{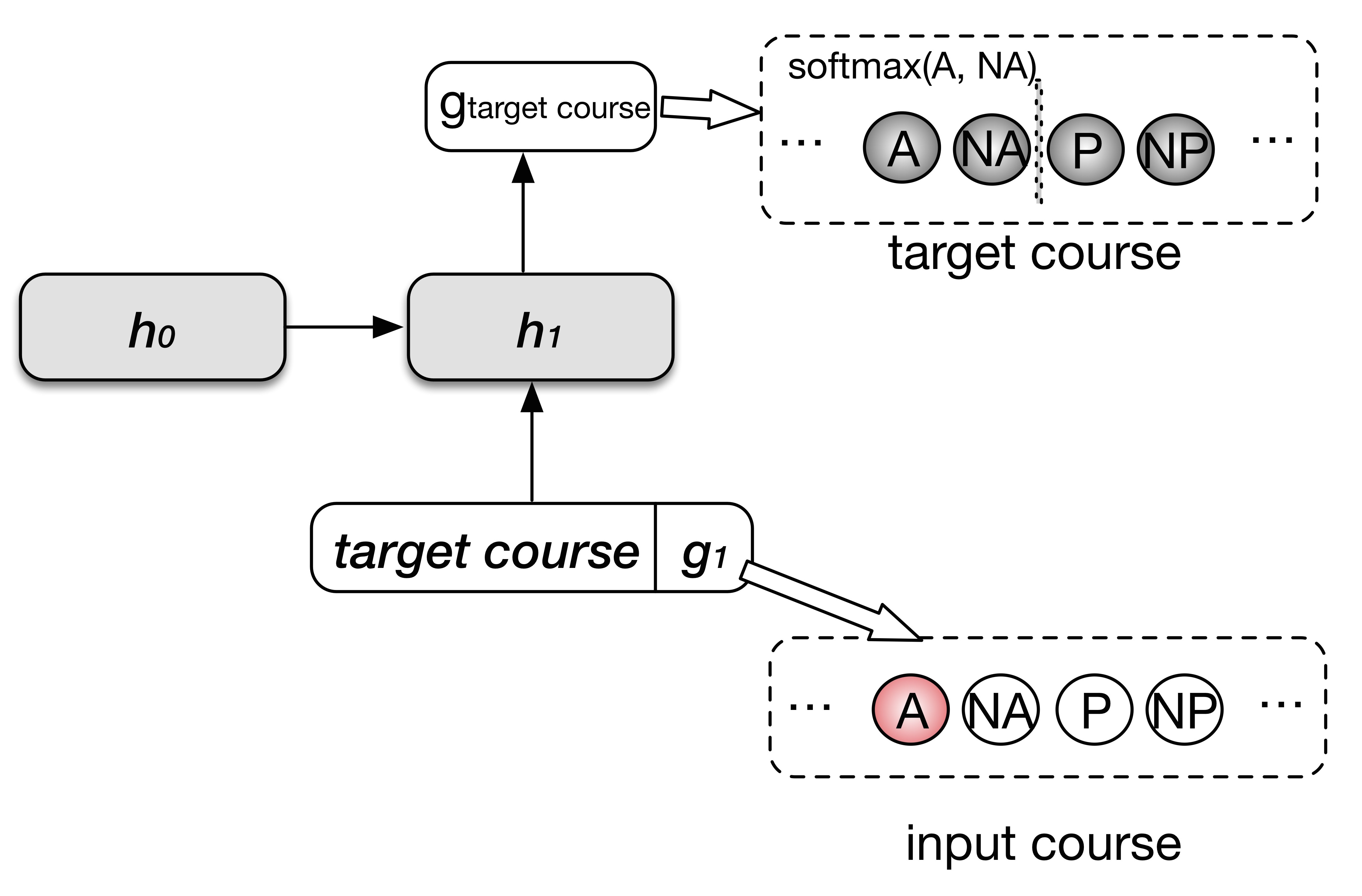}
\caption{Prerequisite course prediction model illustration }
\label{prereqs}
\end{figure}

Note that, for this evaluation, only one time slice input of the grade prediction trained LSTM is needed. The illustrative structure for predicting the prerequisite course for a target course of \textit{Model 2} is shown in Figure \ref{prereqs}. With a student specified grade threshold of A as an example, the steps are\footnote{The steps are the same for the grade threshold of B}\footnote{For \textit{Model 1} and \textit{Model 3}, the steps are the same but only differ in the non-grade part of the LSTM input.}:

\begin{enumerate}

    \item Set $c_1$ to be a one-hot vector with only the position for the target course to be 1 and others positions to be 0.
    \item Iterate $g_1$ over all the courses with only one-hot embedded in the `above or equal to grade A' position for that course, and feed the input, which is a concatenation of $c_1$ and $g_1$, to the learned model described in section 5.
    \item Calculate the predict probability of getting an above or equal to A grade for the target course for each input by $P_{target}^A = \frac{e^{s_{target}^A}}{(e^{s_{target}^A}+e^{s_{target}^{NA}})}$.
    \item Rank $P_{target}^A$ and select the top 10 \footnote{The reason we selected top 10 candidate prerequisite courses is that the largest number of prerequisite courses a target course has in our validation set is 8. So 10 is a proper number to ensure the list may be able to cover all the eight prerequisite courses.} input courses with regard to the value of its output of $P_{target}^A$.
\end{enumerate}

\begin{table}
  \centering
  \caption{Sample of data from prerequisite pairs set}
    \begin{tabular}{l|l}
    \toprule
    \textbf{prerequisite course} & \textbf{target course} \\
    \midrule
    Computer Science 188 & Computer Science 189\\
    
    Mathematics 53 & Computer Science 189\\
    
    Statistics 5 & Computer Science 266\\
    
    Chemistry 3A & Chemistry 3B \\

    Economics 1 & Economics 100B\\
    \bottomrule
    \end{tabular}%
  \label{pairs}%
\end{table}%

We used a set of 2,300 prerequisite course pairs, provided by the UC Berkeley Office of the Registrar, which contains 1,215 target courses, as a source of ground truth to serve as a validation set to test our assertion that the model encodes such relationships and others like them. The basic structure of the prerequisite pairs set is shown in Table \ref{pairs}. Given the over 9,000 course vocabulary of our models, this inference is significantly more challenging than the 69 exercise vocabulary used to construct the prerequisite graph inference in Deep Knowledge Tracing applied to tutoring systems \cite{piech2015deep}. To aid this inference, we applied common sense constraints leveraging existing domain knowledge. We extracted the department information of the prerequisite courses for other courses in the same department as the target course as a filter before step (2). For example, assuming that Table \ref{pairs} shows all of the prerequisite pairs in the university, when we evaluate on \textit{Computer Science 189} as the target course, we only consider candidate prerequisite courses from its own department (i.e., Computer Science) and departments which host the prerequisite courses for the other computer science courses (e.g., \textit{Computer Science 266}). In this case, the candidate departments for prediction are computer science and statistics given that \textit{Statistics 5} is the prerequisite course for \textit{Computer Science 266}. The second filter we added before step (2) is to filter out higher level courses by course number. At UC Berkeley, three levels of courses are provided based on intended year of study, in which the first level courses include courses with numbers lower than 100 (intended for first and second year students), the second level courses include courses with numbers between 100 and 199 (intended for 3rd and 4th year students), and the third level courses include courses with numbers 200 and above (graduate level courses). This filter represents the common sense constraint that a higher difficulty level course should not act as the preparation course for a lower difficulty level course. Since the registrar's list is treated as suggested prerequisites, not enforced by the enrollment system, it is not a highly maintained list, and other alternative prerequisite courses can be expected to exist. The proposed goal-based recommendation method would be ideally able to suggest preparatory courses for any target courses, including preparation courses that may not be in the maintained university list of prerequisites, but may nevertheless be valid. Therefore, this evaluation may represent the lower-bound of prerequisite course retrieval.

\subsection{Results}
Given that a target (post-requisite) course may have several prerequisite courses, We evaluated on the prerequisite course pairs set by calculating:

\begin{enumerate}
\item {the overall accuracy of predicting the correct prerequisite of each of the prerequisites pairs: 
 \[ \frac{\textit{\# correctly predicted prerequisites pairs}}{\textit{\# total prerequisites pairs}} \]
}

 \item {the overall accuracy of predicting at least one prerequisite for each of the post-requisite courses listed in the validation set:
 \begin{equation*}
    \frac{\textit{\# target courses with at least one prerequisite course correctly predicted}}{\textit{\# total target courses}}
\end{equation*}
}
 
\end{enumerate}

The results are shown in Table \ref{tab-pre} and suggest similarity in performance among all models. They show that close to one third of preexisting prerequisites are recovered by all models. For around 44\% of post-requisite courses (courses with at least one prerequisite course), at least one prerequisite course was successfully identified among the top 10 inferred candidates. This result suggests that if these instructor-defined prerequisites are the only courses that could cover the required material, then our prerequisite courses prediction will fail to surface useful recommendations for a little over half of possible candidate courses. Because this list of prerequisites is not expected to be comprehensive, we proceed to a third validation to observe which courses students were choosing to take the semester before a difficult course, assuming that some of their selections may have been made in preparation for the difficult course.

\begin{table}
  \centering
  \caption{Evaluation of prerequisite courses prediction}
    \begin{tabular}{c|c|c|c}
    \toprule
    \multicolumn{2}{c|}{\textbf{model}} & \multicolumn{2}{c}{\textbf{accuracy}} \\
    \midrule
    \textbf{threshold} & \textbf{model} & \textbf{pairs} & \textbf{target courses} \\
    \midrule
    B     & Model 2 & 30.48  & 44.86  \\
    \midrule
    B     & Model 3 & 29.61  & 43.54  \\
    \midrule
    A     & Model 2 & 29.72  & 43.77  \\
    \midrule
    A     & Model 3 & 30.08  & 45.61  \\
    \bottomrule
    \end{tabular}%
  \label{tab-pre}%
  \vspace{-2mm}
\end{table}%

\section{student goal-based course selection prediction}
In the grade prediction validation presented in section 5, the A threshold models performed considerably above baseline, while the B threshold models only slightly outperformed baseline. In spite of this, the course embeddings still encoded essential prerequisite relationships as are justified in section 6, key to the goal-based recommendation in this section. The -goal- in this work refers to a student's desired grade on a target course. In addition, personalization is brought to bear in this section, using a student's personal course history to produce the types of recommendations the framework would make in a real-world setting. This personalization allows for a different set of courses to be tailored to students based on their various enrollment histories, discrepancies in the grasp of course knowledge, and different majors. For example, different prerequisite courses may be tailored to a Global Studies major than a Mechanical Engineering major in preparation for a course on machine learning. To achieve this, we propose the goal-based recommendation in this section to generate personalized suggestions to students by means of the same learned grade prediction models described in section 5 which implicitly model students' developing hidden performance aptitude, or "knowledge" states, across semesters.

\subsection{Personalized Goal-based Prerequisite Course Recommendation Framework}

The goal-based recommendation task can be defined as a response to a student's query, ``To perform well on a target course of interest, which course shall I take the semester before, given my course enrollment and grade history?" The framework for generating the recommended courses for the model is shown in Figure \ref{goal}, which also leverages the same grade prediction LSTM models we have been testing in the two previous evaluations. The framework considers a student's enrollment histories, enrolled courses' grades and a grade goal (A or B) for a target course to generate recommendations for the previous semester that maximizes the predicted probability of a student attaining that goal. 

Assuming that a student has course enrollment histories for $t-1$ semesters, and hopes to perform well (set A as the threshold) in a target course in semester $t+1$. The steps for generating recommended courses for the target course in semester $t$ are:

\begin{figure}
\includegraphics[height=1.4in, width=3.5in]{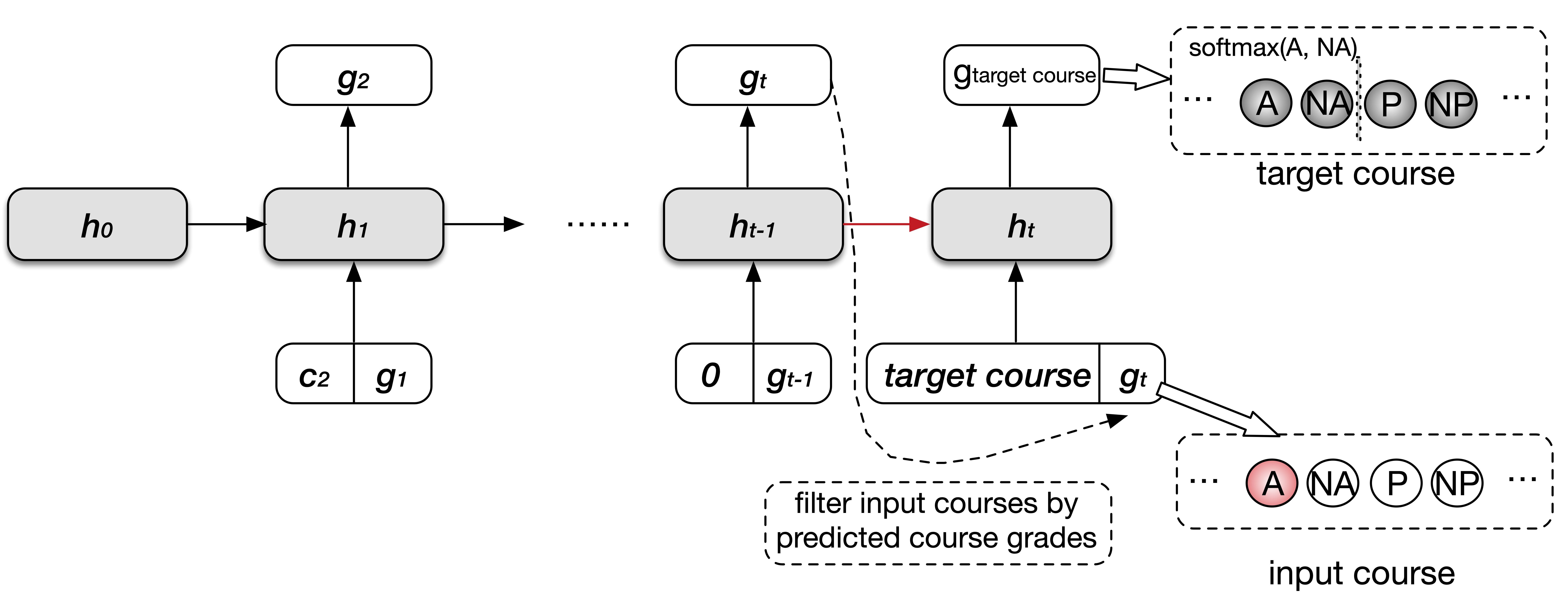}
\caption{Goal-based recommendation model illustration}
\label{goal}
\end{figure}

\begin{enumerate}
    \item Input the student's enrollment histories and grade histories to the model and then retrieve the value for $h_{t-1}$ and $g_{t-1}$. Note that $c_t$ should be set to \textbf{0} for the $t-1$-th input of the model because it will be unknown in the real recommendation scenario.
    
    \item Iterate $g_t$ over all the courses with a one-hot representation of the grade A position for that course and give $g_t$, $h_{t-1}$ and $c_{t+1} = target$ $course$ to the hidden layer.
    
    \item Calculate the predicted probability of getting an A for the target course for each input by $P_{target}^A = \frac{e^{s_{target}^A}}{(e^{s_{target}^A}+e^{s_{target}^{NA}})}$.
    \item Rank $P_{target}^A$ and select the top 10 input courses with regard to the value of its output of $P_{target}^A$.
\end{enumerate}

In order to leverage domain knowledge to limit the number of courses for enumeration without reducing the rigorousness of evaluation, we applied several filters to the enumerated input courses before step (2), which are listed as follows, with the first two filters being similar to those we employed to prerequisite course prediction. 
\begin{enumerate}
    \item We only consider courses in target course's own department and the departments which host the prerequisite courses for the courses in the same department as the target course.
    \item Filter out the higher level courses by the course number.
    \item Filter out courses which are unavailable in semester $t$.
    \item Filter out courses the student has already taken. 
    \item Filter out the target course.
    \item Only consider the courses with predicted grades higher than the threshold (A in Figure \ref{goal}), because the recommended course should also be within the student's zone of proximal development.
\end{enumerate}

We selected 10 historically difficult courses from different departments across STEM and non-STEM disciplines as the target and set  Fall 2016 as the target semester. Because sgnificantly fewer students take courses in summer semesters, we considered Spring 2016 as the semester for recommendation. The test set for this goal-based recommendation validation consists of the students with enrollment histories in Fall 2016 and at least two semester before Fall 2016 but without enrollment histories in Summer 2016.   

We considered all students who took the target courses in Fall 2016. Our model was considered to have made a successful set of recommendations if at least one of the 10 would-be suggested courses, recommended for 2016 Spring, matched a student's actual enrollment in that semester. The overall accuracy was calculated by the number of students we have correct predictions for over the number of total students. In addition, we assume that students who performed well on the target course and students who did not may show differences in the courses they chose to take in the previous semester. Students who performed well on the target course may show more preparation in their enrollment histories that lay solid foundation for the target course. We hypothesize that the recommendation accuracy on high performance students may therefore be higher than that on poorer performance students and that this difference is support, but not proof, for the reasonableness of preparation course recommendations being made. Hence, we report the accuracy on both high performance students and poorer performance students with respect to the target by setting the threshold grade (A or B) to distinguish them.

\subsection{Results}
\begin{figure*}
\includegraphics[height=3in, width=7.2in]{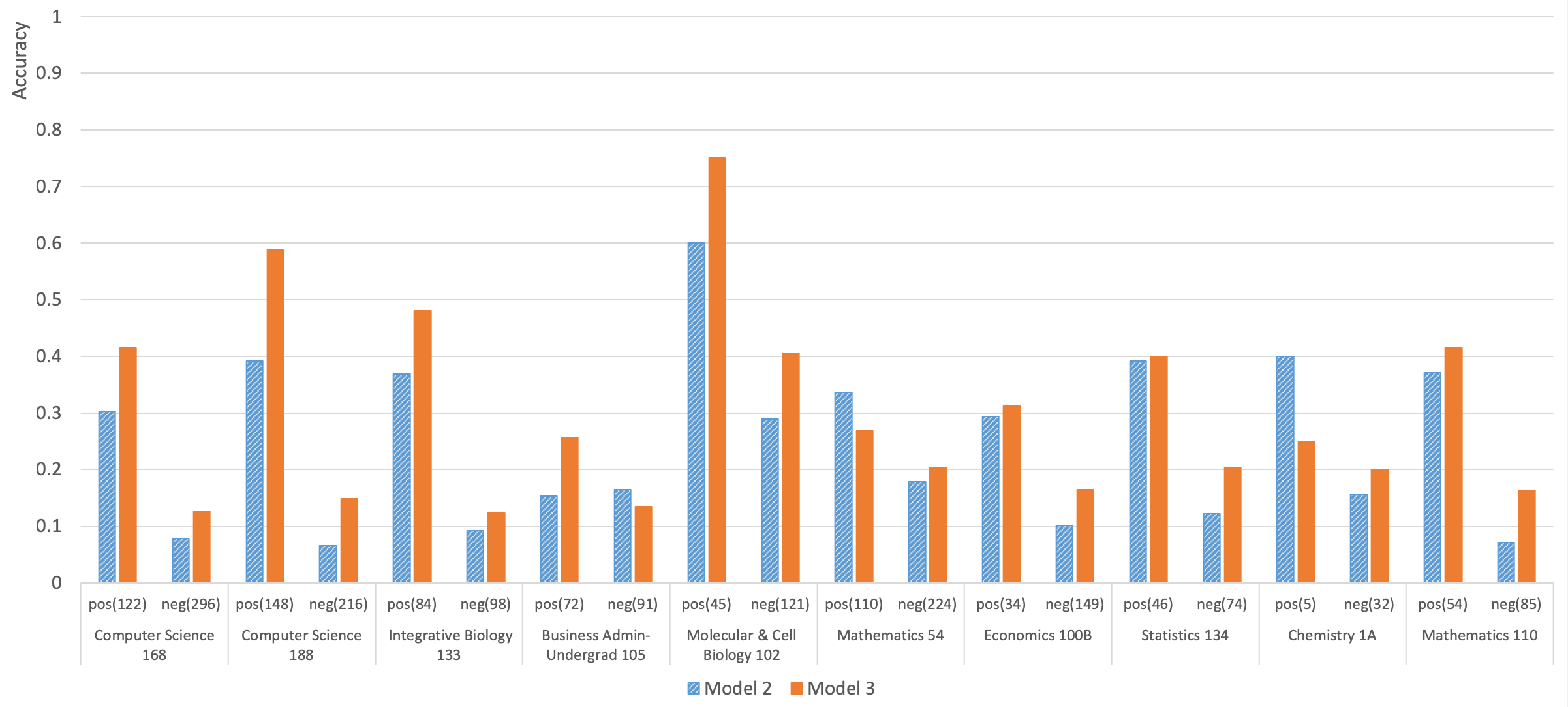}
\caption{Goal-based recommendation evaluation (Grade threshold: A)}
\label{goal_A}
\end{figure*}

\begin{figure*}
\includegraphics[height=3in, width=7.2in]{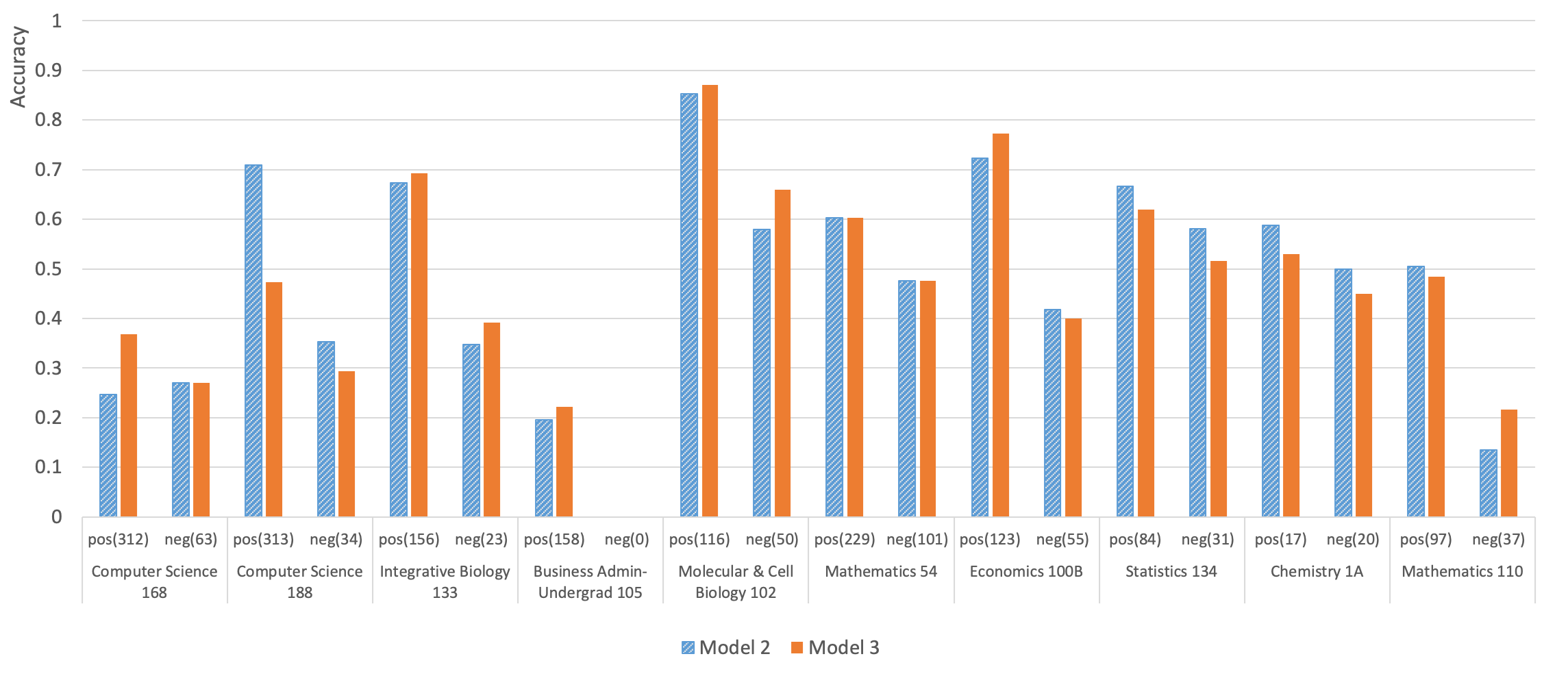}
\caption{Goal-based recommendation evaluation (Grade threshold: B)}
\label{goal_B}
\end{figure*}

The specific results for each course we selected for models with grade threshold set as A and B are shown in Figure \ref{goal_A} and \ref{goal_B}, respectively. Summary results (Table  \ref{tab-sum}) show that students who performed under-threshold on the target course were also far less likely to have taken a course that would have been suggested by the algorithm in the previous semester. For both the A and B threshold scenarios, recommendations matched at least one preparatory course taken by 57\% of students scoring a B or above and 38\% of students scoring an A on the target course. For both threshold models, recommendation matches on students scoring below threshold on the target course were 21 raw percentage points lower in accuracy. 

All three evaluation results are summarized in Table \ref{tab-sum}, combining the course selection evaluation with results from grade prediction (only letter grade) and the prerequisite inference evaluation (using average accuracy). The best performing model for each evaluation is bold. Note that it is the same trained model on each row that is being validated from difference perspectives. The grade prediction validation is closest to the objective function used to train the model. Its direct affect on recommendation is that it is used to apply the filter of not showing students preparatory courses they are not ready for (not predicted to perform above threshold on). In spite of raw accuracy differences and differences between model and baseline accuracies among the different threshold models, all models  performed well in recovering existing prerequisites, maintained by the university.  

\begin{table}
  \centering
  \caption{Summary of all evaluations}
    \begin{threeparttable}
    \begin{tabular}{c|c|c|c|c|c}
    \toprule
    \multicolumn{2}{c|}{\textbf{model}} & \multicolumn{2}{c|}{\textbf{goal-based}} &  \multicolumn{1}{c|}{\textbf{grade pred.}} & \multicolumn{1}{c}{\textbf{pre-req}} \\
    \midrule
    \textbf{thres.} & \textbf{model} & \textbf{pos} & \textbf{pos-neg} &\textbf{F-score} & \textbf{average}\tnote{1} \\
    \midrule
    B     & Model 2 & \textbf{57.65}  &  21.21& 42.01 & 37.67  \\
    \midrule
    B     & Model 3 & 56.34  & 19.33 & 40.21 &36.58\\
    \midrule
    A     & Model 2 & 36.09  & \textbf{22.93} & \textbf{60.24} & 36.74\\
    \midrule
    A     & Model 3 & 41.36  & 22.61 & 58.86 &  \textbf{37.84}\\
    \bottomrule
    \end{tabular}%
    \begin{tablenotes}
        \footnotesize
        \item[1] Calculated by averaging the third and the fourth accuracy columns of Table \ref{tab-pre}.
      \end{tablenotes}
  \end{threeparttable}
  \label{tab-sum}%
\end{table}%

\subsection{Discussion}
Model 2 with a B threshold scored considerably higher than the same model with an A threshold on the preparation semester course prediction evaluation (57.65\% vs. 36.09\%). One explanation is that students are trying to match course difficulty with their ability (i.e., ZPD) when making their selections \cite{ognjanovic2016using} and that, since the B models have higher grade prediction accuracy than the A models (88.05\% vs. 75.23\%), they are better at filtering out courses that are not a good match. The higher letter grade accuracy of the B model, however, is not very meaningful and not directly comparable to the A model since the B model benefits greatly from having a higher majority class proportion than the A model. In terms of grade prediction F-score, the A model is superior (60.24 for Model 2 A vs 42.01 for Model 2 B). If the grade discriminating power of the B model is not its source of strength in the goal-based evaluation, an alternative explanation is that an A threshold for a preparatory class is too high. 
The similar scores between A and B models on the pre-req inference evaluation suggests that receiving an A does not carry substantially more information about pre-requisite relationships than does receiving a B. Therefore, receiving a B may be, on average, sufficient preparation for post-requisite courses and therefore a better threshold. It is worth noting that grade prediction \textit{outputs} are meant to be an estimate of expected achievement in courses given the current knowledge state of a student and a normative amount of effort. If a student decides to dedicate above average time to a class, they can overachieve the model's estimate and gain more from the class than anticipated. Therefore, the predictions should not be treated as boundaries but rather distances from a student's ZPD, traversable with appropriate effort and support. 
\section{Contributions}
We introduced a novel approach to personalized course prerequisite inference for goal-based recommendation based on adaptations of a recurrent neural network. We validated several model variants against test sets representing the tasks of grade prediction, prerequisite inference, and preparation semester course selection. The model allows students to specify an arbitrary course offered at the university along with the level of achievement they wish to attain (a grade of A or B). The algorithm then tailors 10 candidate preparation courses to consider based on their personal course enrollment histories and grades, their specified target course and achievement level. As this is a causal inference problem and we had only observational data to train the model, we used these three sources of validation of a model trained on grade prediction to help gauge the plausibility of the model performing adequately in the real-world. B target threshold models scored slightly above baseline in the grade prediction task, achieving a high of 88\% accuracy on the binary classification task while the A model scored lower at 75\% but substantially beat out the lower performing majority class baseline of 50\%. The grade prediction performance is important in order to accurately filter out preparatory courses from recommendation that a student may not be ready for and may, themselves, require additional preparations for. Since the preparation recommendations are in essence a personalized inference of prerequisite information, we tested the models' ability to encode this information by validating against a preexisting prerequisite graph kept by the university. On this validation set, at least one prerequisite course was recovered for nearly half of the post-requisite courses in the graph and around one third of the total prerequisite pairs were recovered. The prerequisite graph was not expected to be comprehensive and students may be finding ways to prepare for courses in ways other than consulting the courses in this graph. We therefore used student course selections in the semester before a historically difficult course as a third source of validation. The hypothesis being that students who achieved above threshold in the target difficult course would be more likely to have taken a would-be recommendation of our algorithm than students who performed below threshold. Our full personalized goal-based recommendation pipeline was enacted to conduct this evaluation, utilizing students previous course histories before the recommendation semester. Our results showed that students who performed above threshold on the difficult course were nearly twice as likely to have taken at least one of the ten would-be recommended preparation courses of our algorithm compared to students who performed below the specified performance threshold.

\section{Limitations and Future Work}
Inherent limitations of observational data exist in all applications that wish to infer causal relationships. While real-world evaluation with experimental controls is the gold standard; this would be a very expensive undertaking to realize, given that two semesters would need to elapse in order to make the recommendations direct to students, or as an advising tool, and then observe students' performance on the target course. Furthermore, given the high stakes of such a recommendation, we would want to be highly confident in the algorithm's performance in order to ethnically justify such a real-world evaluation. Therefore, back-testing validations, such as those conducted in this paper, are a necessary first step to reaching the goal of real-world impact.

Additional sources of validation that could be sought before deployment would be instructor ratings of suggested prerequisite courses for their own course given to students with different exemplar course histories, academic adviser ratings of such recommendations, and ratings from students themselves. Although among these three stakeholders, students may be least capable of judging if the content of a course they haven't yet taken would serve as appropriate preparation for a course currently outside their proximal zone of development. 

Methodologically, our recommendation was set up to recommend course preparation for a single semester. The best preparation course may be one that the student is not ready to take, and would be filtered out from recommendation. A sequence of curricular preparation may be more desirable, depending on how distant of courses from their current level they wish to engage with and how far in advance they plan.

There are many other goals students wish to achieve, from the micro to the macro. Future work includes evaluating if these goals, such as preparation for an intended career path, fit well into the modeling paradigm described and what additional augmentations, if any, are needed to aid students in navigating those decision spaces.
\appendix

\begin{acks}
We thank the following University of California, Berkeley staff for their assistance with the anonymized data and for sharing their wisdom with respect to the complexities of the enrollment context; Andrew Eppig, Tom O'Brien, Rebecca Sablo, and Walter Wong. This material is based in part upon work supported by the National Science Foundation under Grant Number 1446641.
\end{acks}

\bibliographystyle{ACM-Reference-Format}
\bibliography{sample-bibliography}

\end{document}